\documentclass[letterpaper, 10 pt, conference]{URL-ieeeconf}
\IEEEoverridecommandlockouts    
\usepackage{graphics}           
\usepackage{times}              
\usepackage{amsmath}            
\usepackage{amssymb}            
\usepackage{graphicx}
\usepackage{algorithm}
\usepackage[noend]{algpseudocode}
\usepackage{booktabs}
\usepackage{color}
\usepackage{multirow}
\usepackage{subcaption}
\usepackage{rotating}
\usepackage{cite}
\definecolor{instructioncolor}{rgb}{.5,.5,.5}

\def\secref#1{Section~\ref{#1}}
\def\figref#1{Fig.~\ref{#1}}
\def\tabref#1{Table~\ref{#1}}
\def\eqref#1{(\ref{#1})}

\def\vsfig{\vspace{-0.3cm}}

\newcommand{\rom}[1]{\uppercase\expandafter{\romannumeral #1\relax}}

\makeatletter
\usepackage{xspace}
\DeclareRobustCommand\onedot{\futurelet\@let@token\@onedot}
\def\@onedot{\ifx\@let@token.\else.\null\fi\xspace}

\makeatother

\usepackage{array}
\newcolumntype{L}[1]{>{\raggedright\let\newline\\\arraybackslash\hspace{0pt}}m{#1}}
\newcolumntype{C}[1]{>{\centering\let\newline\\\arraybackslash\hspace{0pt}}m{#1}}
\newcolumntype{R}[1]{>{\raggedleft\let\newline\\\arraybackslash\hspace{0pt}}m{#1}}

\renewcommand{\b}[1]{\mbox{\boldmath$#1$}}

\newcommand{\dI}{\b I}

\newcommand{\dR}{\b R}
\newcommand{\dS}{\b S}

    \newcommand\dIMUFrame{\mathit{I}}
    \newcommand\dLidarFrame{\mathit{L}}
    \newcommand\dBodyFrame{\mathit{B}}
    \newcommand\dWorldFrame{\mathit{W}}

    \newcommand\dIthIMUFrame{\mathit{I}_{i}}
    \newcommand\dJthIMUFrame{\mathit{I}_{j}}
    \newcommand\dKthIMUFrame{\mathit{I}_{k}}

    \newcommand\dJthLidarFrame{\mathit{L}_{j}}
\newcommand\dZero{\mathbf{0}}
        \newcommand\dPose{\mathbf{T}}

        \newcommand\dIthPose{\dPose_{i}}

        \newcommand\dPoseFromFrameIMUToLidar{\dPose_{\dIMUFrame\dLidarFrame}}

        \newcommand\dPoseFromFrameWorldToKthIMU{\dPose_{\dKthIMUFrame}}
        \newcommand\dPoseFromFrameWorldToJthIMU{\dPose_{\dJthIMUFrame}}
        
        \newcommand\dPoseFromFrameKthIMUToJthIMU{\dPose_{\dKthIMUFrame\dJthIMUFrame}}

            \newcommand\dPoseInv{\dPose^{-1}}

            \newcommand\dPoseInvFromFrameWorldToKthIMU{\dPoseInv_{\dKthIMUFrame}}
            
        \newcommand\dPoseNom{\hat{\dPose}}

        \newcommand\dPoseNomFromFrameWorldToJthIMU{\dPoseNom_{\dJthIMUFrame}}

        \newcommand\dPoseNomFromFrameKthIMUToJthIMU{\dPoseNom_{\dKthIMUFrame\dJthIMUFrame}}
        \newcommand\dPoseNomFromFrameWorldToKthIMU{\dPoseNom_{\dKthIMUFrame}}
        \newcommand\dPoseNomInvFromFrameKthIMUToJthIMU{\dPoseNomInv_{\dKthIMUFrame\dJthIMUFrame}}
        
            \newcommand\dPoseNomInv{\dPoseNom^{-1}}

            \newcommand\dPoseNomInvFromFrameWorldToKthIMU{\dPoseNomInv_{\dKthIMUFrame}}
            \newcommand\dPoseNomInvFromFrameKthIMUToIthIMU{\dPoseNomInv_{\dKthIMUFrame\dIthIMUFrame}}
        \newcommand\dPosePert{\boldsymbol{\xi}}

        \newcommand\dPosePertFromFrameWorldToJthIMU{\dPosePert_{\dJthIMUFrame}}
        \newcommand\dPosePertFromFrameWorldToKthIMU{\dPosePert_{\dKthIMUFrame}}
        
        \newcommand\dPosePertFromFrameKthIMUToJthIMU{\dPosePert_{\dKthIMUFrame\dJthIMUFrame}}
        
            \newcommand\dPosePertTrans{\dPosePert^{\top}}

            \newcommand\dPosePertTransFromFrameKthIMUToJthIMU{\dPosePertTrans_{\dKthIMUFrame\dJthIMUFrame}}
            \newcommand\dPosePertPrime{\dPosePert^{\prime}}

            \newcommand\dPosePertPrimeFromFrameWorldToKthIMU{\dPosePertPrime_{\dKthIMUFrame}}

                \newcommand\dPosePertPrimeCurwed{\dPosePert^{\prime\curlywedge}}

                \newcommand\dPosePertPrimeCurwedFromFrameWorldToKthIMU{\dPosePertPrimeCurwed_{\dKthIMUFrame}}
        \newcommand\dPoseCov{\mathbf{\Sigma}}

        \newcommand\dRot{\mathbf{R}}

        \newcommand\dIthRot{\dRot_{i}}

        \newcommand\dRotTrans{\dRot^{\top}}

        \newcommand\dIthRotTrans{\dRotTrans_{i}}

        \newcommand\dTls{\mathbf{t}}

        \newcommand\dIthTls{\dTls_{i}}

        \newcommand\dPt{\mathbf{p}}

        \newcommand\dJthPt{\dPt_{j}}

        \newcommand\dJthPtInJthLidarFrame{{}^{\dJthLidarFrame} \dPt_{j}}
        \newcommand\dJthPtInKthIMUFrame{{}^{\dKthIMUFrame} \dPt_{j}}

            \newcommand\dPtHomo{\dPt^{\mathrm{h}}}

            \newcommand\dJthPtHomoInJthLidarFrame{{}^{\dJthLidarFrame} \dPtHomo_{j}}
            \newcommand\dJthPtHomoInKthIMUFrame{{}^{\dKthIMUFrame} \dPtHomo_{j}}
            
        \newcommand\dPtNom{\hat{\dPt}}

        \newcommand\dJthPtNomInKthIMUFrame{{}^{\dKthIMUFrame} \dPtNom_{j}}
            \newcommand\dPtNomHomo{\dPtNom^{\mathrm{h}}}

            \newcommand\dJthPtNomHomoInKthIMUFrame{{}^{\dKthIMUFrame} \dPtNomHomo_{j}}
            \newcommand\dJthPtNomHomoInJthIMUFrame{{}^{\dJthIMUFrame} \dPtNomHomo_{j}}
            \newcommand\dPtNoiseDil{\mathbf{n}^{\mathrm{d}}}

            \newcommand\dJthPtNoiseDilInKthIMUFrame{{}^{\dKthIMUFrame} \mathbf{n}_{\dJthPt}^{\mathrm{d}}}
            \newcommand\dJthPtNoiseDilInJthIMUFrame{{}^{\dJthIMUFrame} \mathbf{n}_{\dJthPt}^{\mathrm{d}}}
        \newcommand\dState{\mathbf{x}}

        \newcommand\dIthState{\dState_{i}}
        
        \newcommand\dKthState{\dState_{k}}

            \newcommand\dStateTrans{\dState^{\top}}
        \newcommand\dStateNom{\hat{\dState}}

        \newcommand\dIthStateNom{\dStateNom_{i}}
        \newcommand\dKthStateNom{\dStateNom_{k}}
        \newcommand\dStateEst{\hat{\dState}}

        \newcommand\dStateErr{\tilde{\dState}}

        \newcommand\dIthStateErr{\dStateErr_{i}}
        \newcommand\dKthStateErr{\dStateErr_{k}}
        \newcommand\dVel{\mathbf{v}}

        \newcommand\dVelErr{\tilde{\dVel}}

        \newcommand\dVelWorld{ {}^{\dWorldFrame} \dVel}

        \newcommand\dIthVelWorld{\dVelWorld_{i}}

        \newcommand\dVelBody{ {}^{\dBodyFrame} \dVel}
        \newcommand\dIthVelBody{\dVelBody_{i}}

            \newcommand\dVelBodyTrans{\dVelBody^{\top}}

        \newcommand\dTwist{\boldsymbol{\zeta}}

        \newcommand\dIthTwist{\dTwist_{i}}

            \newcommand\dGyroBias{\mathbf{b}^{\dGyro}}

        \newcommand\dIthGyroBias{\dGyroBias_{i}}
            \newcommand\dGyroBiasErr{\tilde{\mathbf{b}}^{\dGyro}}
            \newcommand\dAccBias{\mathbf{b}^{\dAcc}}

        \newcommand\dIthAccBias{\dAccBias_{i}}
            \newcommand\dAccBiasErr{\tilde{\mathbf{b}}^{\dAcc}}
            \newcommand\dGyroBiasNoise{\mathbf{n}_{\dGyroBias}}
            \newcommand\dGyroBiasNoiseTrans{\dGyroBiasNoise^{\top}}
            \newcommand\dAccBiasNoise{\mathbf{n}_{\dAccBias}}
            \newcommand\dAccBiasNoiseTrans{\dAccBiasNoise^{\top}}
        \newcommand\dGyro{\boldsymbol{\omega}}

        \newcommand\dIthGyro{\dGyro_{i}}
        \newcommand\dGyroTrans{\dGyro^{\top}}

        \newcommand\dIthGyroTrans{\dGyroTrans_{i}}
        \newcommand\dGyroNoise{\mathbf{n}_{\dGyro}}
        \newcommand\dGyroNoiseTrans{\dGyroNoise^{\top}}
        \newcommand\dAcc{\mathbf{a}}

        \newcommand\dIthAcc{\dAcc_{i}}
        \newcommand\dAccTrans{\dAcc^{\top}}

        \newcommand\dIthAccTrans{\dAccTrans_{i}}
        \newcommand\dAccNoise{\mathbf{n}_{\dAcc}}
        \newcommand\dAccNoiseTrans{\dAccNoise^{\top}}

\def\dSE3{\mathrm{SE}(3)}
\def\dSO3{\mathrm{SO}(3)}
\def\dS2{\mathbb{S}^2}
\def\dR3{\mathbb{R}^3}

\newcommand\dVecSpace{\mathbb{R}}

\newcommand\dBigExp{\mathrm{Exp}}

\usepackage{url}

\usepackage{titlesec}
\titlespacing{\section}{0pt}{6pt}{4pt}
\titlespacing{\subsection}{0pt}{4pt}{2pt}

\usepackage{xcolor}

\title{\LARGE \bf $\boldsymbol{\mathbf{\dSE3}}$-LIO: Smooth IMU Propagation With Jointly Distributed Poses \\ on $\boldsymbol{\mathbf{\dSE3}}$ Manifold for Accurate and Robust LiDAR-Inertial Odometry}

\author{Gunhee Shin$^{1}$, Seungjae Lee$^{1}$, Jei Kong$^{1}$, Youngwoo Seo$^{2}$ and Hyun Myung$^{1*}$
  \thanks{$^*$Corresponding author: Hyun Myung}
  \thanks{$^{1}$School of Electrical Engineering, KAIST (Korea Advanced Institute of Science and Technology), Daejeon, 34141, South Korea, {\tt\scriptsize \{gunhee$\_$shin, sj98lee, kongj0531, hmyung\}@kaist.ac.kr}}
  \thanks{$^{2}$\raggedright Hanwha Aerospace, 6, Pangyo-ro 319 beon-gil, Bundang-gu, Seongnam-si, Gyeonggi-do, South Korea. {\tt\scriptsize \{youngwoo.seo\}@hanwha.com}}
}
\begin{document}
\maketitle
\thispagestyle{empty}
\pagestyle{empty}

\begin{abstract}
  In estimating odometry accurately, an inertial measurement unit (IMU) is widely used owing to its high-rate measurements, which can be utilized to obtain motion information through IMU propagation.
  In this paper, we address the limitations of existing IMU propagation methods in terms of motion prediction and motion compensation.
  In motion prediction, the existing methods typically represent a 6-DoF pose by separating rotation and translation and propagate them on their respective manifold, so that the rotational variation is not effectively incorporated into translation propagation.
  During motion compensation, the relative transformation between predicted poses is used to compensate motion-induced distortion in other measurements, while inherent errors in the predicted poses introduce uncertainty in the relative transformation.
  To tackle these challenges, we represent and propagate the pose on $\dSE3$ manifold, where propagated translation properly accounts for rotational variation.
  Furthermore, we precisely characterize the relative transformation uncertainty by considering the correlation between predicted poses, and incorporate this uncertainty into the measurement noise during motion compensation.
  To this end, we propose a LiDAR-inertial odometry (LIO), referred to as $\dSE3$-LIO, that integrates the proposed IMU propagation and uncertainty-aware motion compensation (UAMC).
  We validate the effectiveness of $\dSE3$-LIO on diverse datasets.
  Our source code and additional material are available at: \url{https://se3-lio.github.io/}.
\end{abstract}

\section{Introduction}
\label{sec:intro}

Accurate odometry estimation is essential for autonomous navigation\cite{forster2017tro, Mangelson2020tro}.
Among various proprioceptive sensors, inertial measurement unit (IMU) is widely used with other exteroceptive sensors owing to its high-rate measurements. 
These measurements can be utilized to obtain motion information through IMU propagation, which integrates IMU measurements over time based on a kinematic model\cite{forster2017tro,wei2021ral}.

IMU propagation enhances odometry performance in two aspects: $\textit{motion prediction}$ and $\textit{motion compensation}$.
Firstly, it enables motion prediction before the measurements from other modalities, such as LiDAR point clouds and camera images, which are highly informative but available at lower frequencies.
The predicted motion is used as a prior information for the update step in Kalman filter\cite{wei2021ral,wei2022tro,wu2024icra,geneva2020icra} or a factor for pose graph optimization\cite{shan2020iros,qin2018tro,lim2022uv}.
Secondly, it allows motion compensation for motion-induced distortions in other measurements.
Sensors such as LiDAR and event cameras accumulate samples over a certain period of time, during which a robot's continuous motion would introduce distortions to the measurements\cite{chen2023icra,mueggler2018tro}.
These distortions are typically compensated by transforming the samples into a common reference pose.
Such motion compensation is achieved by applying a relative transformation between the reference pose and the pose at each sample time, where both poses are obtained through IMU propagation.

\begin{figure}[t]
    \captionsetup{font=footnotesize}
    \centering
    \includegraphics[width=0.48\textwidth]{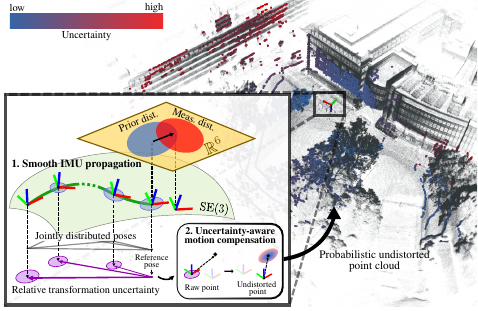}
    \caption{
        Overview of the proposed IMU propagation and LiDAR–inertial odometry (LIO), referred to as $\dSE3$-LIO. 
        The IMU propagation is performed on $\dSE3$ manifold to yield a smooth trajectory and accurate prior.
        The joint distribution of predicted poses is derived to characterize the relative transformation uncertainty by considering their correlation, which is incorporated into the measurement noise during motion compensation to construct a probabilistic undistorted point cloud.
        Finally, $\dSE3$-LIO optimizes both the prior and the measurement distribution, improving the performance of odometry estimation.
    }
    \label{fig:main_figure}
    \vsfig
\end{figure}

While IMU propagation has been extensively studied\cite{forster2017tro,chen2023icra,wu2024icra}, several challenges remain in motion prediction and motion compensation.
For motion prediction, conventional methods typically represent a 6-DoF pose by separating rotation and translation, and propagate them on their respective manifolds\cite{forster2017tro}.
Consequently, rotational variation is not effectively incorporated into translation propagation, resulting in inaccurate motion prediction.
For motion compensation, the predicted poses obtained through IMU propagation inherently contain errors due to sensor noise and numerical integration\cite{brossard2022tro}, introducing uncertainty in the relative transformation used for motion compensation.
This uncertainty should be addressed properly during motion compensation, especially for measurements at distant ranges, such as LiDAR point clouds, where even small rotational changes could introduce substantial perturbations on the measurements\cite{chen2023icra,jung2023ral}.
However, characterizing its uncertainty remains challenging, because it cannot be directly measured from IMU propagation\cite{Mangelson2020tro}.


To address these challenges, we propose a smooth IMU propagation with jointly distributed poses on SE(3) manifold.
The proposed method represents and propagates the pose on SE(3) manifold, where the propagated translation properly accounts for rotational variation, leading to more accurate motion prediction.
Furthermore, we derive the joint distribution of the predicted poses, which enables precise characterization of relative transformation uncertainty by considering their correlation.
This uncertainty is incorporated into the measurement noise during motion compensation to obtain a probabilistic undistorted measurement, and we refer to this process as uncertainty-aware motion compensation (UAMC).
To this end, we present an error-state Kalman filter-based LiDAR–inertial odometry (LIO) framework, termed $\textit{SE(3)-LIO}$, which integrates the proposed IMU propagation and UAMC.
It optimizes a prior from the proposed IMU propagation together with a measurement distribution constructed through UAMC.
We validate the effectiveness of $\dSE3$-LIO with data acquired in diverse scenarios, including a drone under aggressive motion and a ground vehicle operating in large-scale rough terrain, and the results demonstrate improved accuracy and robustness of $\dSE3$-LIO compared with state-of-the-art LIO methods.
The main contributions of this study are as follows:
\begin{enumerate}
    \item We propose a smooth IMU propagation on $\dSE3$ manifold that enables accurate motion prediction by incorporating rotational variation into translation propagation.
    \item We introduce an uncertainty-aware motion compensation (UAMC) approach that constructs a probabilistic undistorted measurement.
    \item We present an accurate and robust LiDAR-inertial odometry (LIO) framework, termed $\dSE3$-LIO, which integrates the proposed IMU propagation and UAMC.
    \item Experimental results in diverse scenarios demonstrate improved accuracy and robustness of $\dSE3$-LIO compared with state-of-the-art LIO methods.
\end{enumerate}
\section{Related Works}
\label{sec:related_works}

\subsection{IMU Propagation}

Extensive IMU propagation methods have been developed based on the preintegration theory~\cite{forster2017tro}, which properly addresses a rotational motion on $\mathrm{SO}(3)$ manifold.
The theory introduces a kinematic model~\cite{murray1994manipulation} that describes the evolution of a 6-DoF by rotation and translation separately.
Building upon this theory, LIO-EKF\cite{wu2024icra} incorporated  high-order terms during discretization of the kinematic model to capture the specific robot motion.
Brossard et al.\cite{brossard2022tro} further extended the theory to $\mathrm{SE}_{2}(3)$ manifold, which jointly represents rotation, translation, and velocity in a single Lie group.
While these methods attempt to improve motion description, they still have limitations in addressing rotational variations during translation propagation, which results in inaccurate motion prediction under the constant global acceleration assumption\cite{forster2017tro,brossard2022tro}.
In parallel, continuous-time representation approaches have been introduced to overcome the limitations of discrete-time models \cite{mueggler2018tro}.
Specifically, B-spline interpolation with the predicted poses as control points has been employed to model smooth trajectories, demonstrating improved performance in environments with curved trajectories~\cite{ramezani2022arxiv,nguyen2024ral,jung2023ral}.
Despite these advancements, they introduce additional complexity and computational overhead, limiting their feasibility in real-time applications.

Moreover, motion compensation is performed using the relative transformation between a reference pose and the pose at each sample time (sample pose).
A large body of works, including LIO-SAM\cite{shan2020iros} and FAST-LIO\cite{wei2021ral}, estimate the sample pose by linearly interpolating between two consecutive predicted poses.
While this approach achieves effective and real-time performance, it operates in a discrete-time manner, which introduces errors particularly in highly dynamic environments.
To address this limitation, recent works~\cite{ramezani2022arxiv,nguyen2024ral} introduced continuous-time models and extracted the sample pose from a smooth trajectory.
Furthermore, DLIO\cite{chen2023icra} introduces a coarse-to-fine IMU propagation approach that incorporates linear jerk and angular acceleration to refine the estimation of the sample pose between predicted poses.
Despite these advances, the predicted poses inherently contain errors due to sensor noise and numerical integration\cite{brossard2022tro}, which consequently introduce uncertainty in the relative transformation for motion compensation.
MA-LIO\cite{jung2023ral} incorporates such uncertainty into measurement noise during motion compensation.
However, it overestimates the uncertainty by neglecting the correlation between predicted poses.
Accurately characterizing the relative transformation uncertainty remains challenging, as it cannot be directly estimated from IMU propagation.

\subsection{LiDAR-Inertial Odometry}
LiDAR–inertial odometry (LIO) has been widely utilized in various robotic applications due to its accuracy and real-time performance~\cite{lee2024isr}.
Among various approaches, Kalman filter methods are preferred for their lightweight formulation and computational efficiency\cite{qin2020icra,lim2023ur}.
FAST-LIO~\cite{wei2021ral} and FAST-LIO2~\cite{wei2022tro} introduced an efficient error-state Kalman filter (ESKF) framework and an incremental kd-tree structure for map management, achieving reliable performance in diverse scenarios.
Building upon this, VoxelMap~\cite{Yuan2022ral} presents an uncertainty-aware ESKF framework that incorporates the prior uncertainty from IMU propagation into measurement noise, which is subsequently propagated to the residual in the measurement distribution.
It also employs an adaptive voxel map that efficiently manages planar features with their associated uncertainty, organized using a hash table and octree structure.
In this work, we build upon the ESKF framework presented in~\cite{wei2021ral,jung2023ral} and enhance odometry performance by integrating our proposed smooth IMU propagation and UAMC.
\section{Smooth IMU Propagation With Jointly Distributed Poses on SE(3) Manifold}
\label{sec:methodolgies}
\subsection{IMU Propagation on SE(3) Manifold}
\label{subsec:imu_propagation_on_SE3_manifold}

A typical IMU returns angular velocity $\dGyro^{m}$ and linear acceleration $\dAcc^{m}$ as measurements, which are affected by biases and noise.
The biases, $\dGyroBias$ and $\dAccBias$, follow a random walk process that varies slowly over time, while the noise, $\dGyroNoise$ and $\dAccNoise$, are modeled as zero-mean white Gaussian distributions\cite{forster2015rss,wei2021ral}.
True IMU value, $\dGyro$ and $\dAcc$, can be expressed as:
\begin{equation} 
    \dGyro^{m} = \dGyro + \dGyroBias + \dGyroNoise, \quad
    \dAcc^{m} = \dAcc + \dAccBias + \dAccNoise.
    \label{eq:imu_measurement}
\end{equation}

Conventional IMU propagation methods typically define the state vector to include following components:
\begin{equation} 
    \dState  \doteq \begin{bmatrix} \dRot \quad \dTls \quad \dVelWorld \quad \dGyroBias \quad \dAccBias \end{bmatrix} \in \dSO3 \times \dVecSpace^{12}
    \label{eq:vanilla_state}
\end{equation}
where the 6-DoF pose is represented by separating rotation and translation, $\dRot$ and $\dTls$, respectively, and
$\dVelWorld$ is the velocity in world frame\cite{forster2017tro,chen2023icra,wu2024icra}.
The state evolves according to the following continuous-time kinematic model:
\begin{equation} 
    \begin{gathered}
        \dot{\dRot} = \dRot (\dGyro)^\wedge, \quad \dot{\dTls} = \dVelWorld, \quad {}^{\dWorldFrame} \dot{\dVel} = \dRot \dAcc + \mathbf{g}, \\
        \dot{\mathbf{b}}^{\dGyro} = \dGyroBiasNoise, \quad \dot{\mathbf{b}}^{\dAcc} = \dAccBiasNoise
    \end{gathered}
    \label{eq:vanilla_continuous_time_kinematic_model}
\end{equation}
where $\dot{(\cdot)}$ denotes the time derivative, $(\cdot)^\wedge$ denotes the skew-symmetric matrix of a vector, and $\mathbf{g}$ is the gravity vector.
By discretizing \eqref{eq:vanilla_continuous_time_kinematic_model}, we obtain the following model:
\begin{equation} 
    \dState_{i+1} = \textbf{f}(\dState_i, \mathbf{u}_i, \mathbf{w}, \Delta t) = 
    \begin{bmatrix} \dIthRot \dBigExp ( \dIthGyro \Delta t )  \\
        \dIthTls + \dIthVelWorld \Delta t \\
        \dIthVelWorld + (\dIthRot \dIthAcc + \mathbf{g}) \Delta t \\
        \dIthGyroBias + \dGyroBiasNoise \\
        \dIthAccBias + \dAccBiasNoise
    \end{bmatrix}
    \label{eq:vanilla_discretized_kinematic_model}
\end{equation}
where $\mathbf{u}_i = [ \dIthGyroTrans \; \dIthAccTrans]^{\top}$ is an IMU input, $\mathbf{w} = [\dGyroNoiseTrans \; \dAccNoiseTrans \; \dGyroBiasNoiseTrans \; \dAccBiasNoiseTrans]^{\top}$ is a process noise term, and $\Delta t$ is the time difference between two consecutive IMU inputs.
For simplicity, $\dIthRot$ in the velocity term is assumed constant during $\Delta t$ under the constant global acceleration assumption \cite{forster2017tro, brossard2022tro}.
Notably, rotation and translation are propagated on their respective manifolds, such that rotational variation is not effectively incorporated into translation propagation.
As robot motion involves simultaneous rotation and translation, neglecting the rotation variation can significantly contribute to pose error, especially under aggressive motion.
The error accumulates through successive IMU propagation, leading to an unreliable prior that degrades the stability of system.

\begin{figure}[t]
    \centering
    \includegraphics[width=0.48\textwidth]{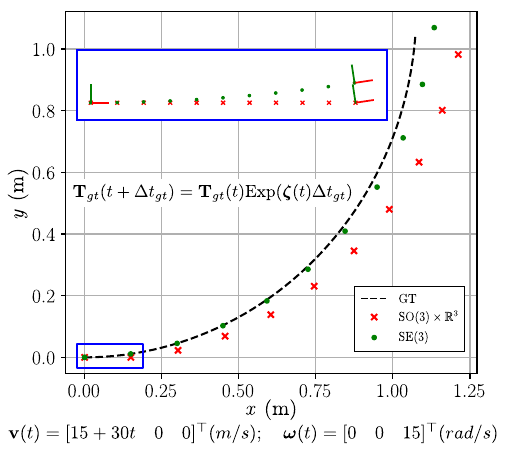}
    \captionsetup{font=footnotesize}
    \caption{
    Illustration of the difference between the conventional\cite{forster2017tro} and the proposed IMU propagation method, which are denoted as $\mathrm{SO}(3) \times \mathbb{R}^3$ and $\mathrm{SE}(3)$, respectively.
    The twist, $\dTwist(t)\doteq[\dVel(t) \; \dGyro(t)]$, is defined as a variable to simulate aggressive motion\cite{foehn2021sr} and generate the ground-truth trajectory with a sufficiently small time step, $\Delta t_{gt}=10^{-4}$ (black dashed line).
    Both methods are performed with 10 IMU inputs, each computed from adjacent twist values, and an interval of $\Delta t = 10^{-2}$.
    The outputs of the proposed method (green dots) yield a smoother trajectory and follow the ground truth (black dashed line) more closely compared with the outputs of the conventional method (red crosses).
    The highlighted box shows that the proposed method properly addresses rotational changes while propagating translation, whereas the conventional method does not.
    } 
    \label{fig:PSE3}
    \vsfig
\end{figure}

To address this issue, we propose a smooth IMU propagation method that evolves poses on SE(3) manifold.
We first represent the 6-DoF pose on SE(3) manifold using a transformation matrix $\dPose$ and define the state vector as:
\begin{equation} 
    \dState \doteq \begin{bmatrix} \dPose \quad \dVelBody \quad \dGyroBias \quad \dAccBias \end{bmatrix} \in \dSE3 \times \dVecSpace^{9}
    \label{eq:our_state_representation}
\end{equation}
where $\dVelBody$ is the velocity in the body frame.
Then, we introduce a continuous-time kinematic model with rigid body motion model\cite{murray1994manipulation} as:
\begin{equation} 
    \dot{\dPose} = \dPose ( \dTwist)^\wedge, \quad {}^{\dWorldFrame} \dot{\dVel}  = \dot{(\dRot \dVelBody)} = \dRot \dAcc + \mathbf{g}
    \label{eq:our_continuous_time_kinematic_model}
\end{equation}
where $\dTwist = \begin{bmatrix} \dVelBodyTrans & \dGyroTrans \end{bmatrix}^{\top} \in \dVecSpace^{6}$ is the twist in the body frame and the biases evolve as in \eqref{eq:vanilla_continuous_time_kinematic_model}.
By discretizing \eqref{eq:our_continuous_time_kinematic_model}, we obtain:
\begin{equation} 
    \begin{aligned}
    \dState_{i+1} &= f(\dState_i, \mathbf{u}_i, \mathbf{w}, \Delta t) \\
    &=
    \begin{bmatrix} \dIthPose \dBigExp ( \dIthTwist \Delta t )  \\
        \dBigExp ( -\dIthGyro \Delta t )\{ \dIthVelBody +(\dIthAcc + \dIthRotTrans \mathbf{g}) \Delta t\} \\
        \dIthGyroBias + \dGyroBiasNoise \\
        \dIthAccBias + \dAccBiasNoise
    \end{bmatrix}.
    \end{aligned}
    \label{eq:our_discretized_kinematic_model}
\end{equation}
The pose is propagated on $\dSE3$ manifold, where the propagated translation properly accounts for rotational variation as:
\begin{equation} 
    \begin{aligned}
        \dPose_{i+1} &= \dIthPose \dBigExp(\dIthTwist \Delta t) \\
        &= \begin{bmatrix} \dIthRot & \dIthTls \\ \dZero & 1 \end{bmatrix} 
        \begin{bmatrix} \dBigExp(\dIthGyro \Delta t) & \mathbf{J}_{l}(\dIthGyro \Delta t) \dIthVelBody \Delta t \\ \dZero & 1 \end{bmatrix}
    \end{aligned}
    \label{eq:our_pose_propagation}
\end{equation}
where $\mathbf{J}_{l}$ denotes the left Jacobian of the exponential map on SO(3) manifold\cite{forster2017tro}.
The propagated translation can be rewritten as:
\begin{equation} 
    \begin{aligned}
        \dTls_{i+1} &= \dIthTls + \dIthRot \mathbf{J}_{l}(\dIthGyro \Delta t) \dIthVelBody \Delta t \\
        &= \dIthTls + \mathbf{J}_{l}(\dIthRot \dIthGyro \Delta t) \dIthVelWorld \Delta t 
    \end{aligned}
    \label{eq:our_translation_propagation}
\end{equation}
where $\dIthRot \mathbf{J}_{l}(\dIthGyro) \dIthRotTrans = \mathbf{J}_{l}(\dIthRot \dIthGyro)$ and $\dIthVelWorld = \dIthRot \dIthVelBody$.
Comparing \eqref{eq:our_translation_propagation} with the conventional method~\eqref{eq:vanilla_discretized_kinematic_model} reveals that
their discrepancy increases with larger twist (dynamic motion) and longer intervals (low-frequency IMU) as illustrated in \figref{fig:PSE3}.
To provide a reliable prior through IMU propagation, both pose and its uncertainty must be propagated accurately\cite{forster2017tro, brossard2022tro}.
We characterize the state uncertainty using the error state, $\dStateErr$, which follows a zero-mean Gaussian distribution as:
\begin{equation} 
    \dStateErr \doteq \begin{bmatrix}
        \dPosePert & \dVelErr & \dGyroBiasErr & \dAccBiasErr
    \end{bmatrix}\in \mathbb{R}^{15}
    \label{eq:error_state_representation}
\end{equation}
where $\dPosePert$ is the perturbation represented on the tangent space of SE(3) manifold and the others are additive perturbations on Euclidean space\cite{wei2021ral}.
The true state $\dState$ is related to the noise-free nominal state $\dStateEst$ and an error state $\dStateErr$ as:
\begin{equation} 
        \dState = \dStateEst \boxplus \dStateErr, \quad \dStateErr = \dState \boxminus \dStateEst
    \label{eq:relationship_between_true_nominal_and_error_state}
\end{equation}
where $\boxplus$ and $\boxminus$ are operators that map between a local neighborhood on manifold and its tangent space \cite{wei2021ral}.
Accordingly, the error state dynamic model can be formulated as:
\begin{equation} 
    \begin{aligned}
    \dStateErr_{i+1} = \dState_{i+1} \boxminus \dStateEst_{i+1} &= f(\dIthState, \mathbf{u}_{i}, \mathbf{w}, \Delta t) \boxminus f(\dIthStateNom, \mathbf{u}_{i}, \mathbf{0}, \Delta t) \\
    &\approx \mathbf{F}_{\dIthStateErr} \dIthStateErr + \mathbf{F}_{\mathbf{w}_{i}} \mathbf{w}
    \end{aligned}
    \label{eq:error_state_dynamic_model}
\end{equation}
where $\mathbf{F}_{\dIthStateErr}$ and $\mathbf{F}_{\mathbf{w}_{i}}$ are the Jacobians of $\dStateErr_{i+1}$ with respect to $\dIthStateErr$ and $\mathbf{w}$, respectively, evaluated at $\dIthStateErr = \dZero$ and $\mathbf{w} = \dZero$\cite{wei2021ral}.
Finally, the state covariance is obtained by taking the expectation of the outer product of \eqref{eq:error_state_dynamic_model} as follows:
\begin{equation} 
    \hat{\mathbf{P}}_{i+1} = \dBigExp \left( \dStateErr_{i+1} \dStateErr_{i+1}^{\top} \right) = \mathbf{F}_{\dIthStateErr} \hat{\mathbf{P}}_{i} \mathbf{F}_{\dIthStateErr}^{\top} + \mathbf{F}_{\mathbf{w}_{i}} \mathbf{Q}_{\mathbf{w}} \mathbf{F}_{\mathbf{w}_{i}}^{\top}
    \label{eq:covariance_representation}
\end{equation}
where $\mathbf{Q}_{\mathbf{w}}$ is the covariance of the process noise.
\subsection{Uncertainty-aware Motion Compensation (UAMC)}
\label{subsec:UAMC}

Consider a measurement that accumulates samples over time, such as LiDAR points and event camera data. 
While the set of samples is treated as a single measurement, each sample is acquired at a different location, resulting in motion-induced distortion. 
To correct this distortion, we transform all samples into a common reference pose.

\begin{figure}[t]
    \centering
    \captionsetup{font=footnotesize}
    \includegraphics[width=0.48\textwidth]{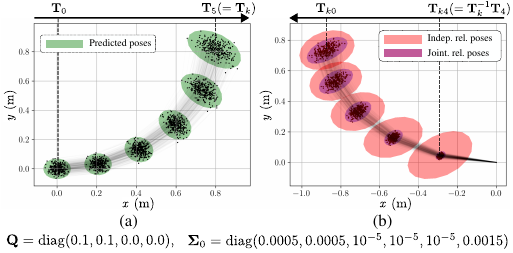}
    \caption{
    Uncertainty characterization of predicted poses from IMU propagation and relative transformations for motion compensation.
    (a) IMU propagation is performed using 100 IMU inputs computed same as \figref{fig:PSE3},
    with a process noise covariance $\mathbf{Q}$.
    Initial pose is set to origin with covariance $\dPoseCov_0$.
    (b) 
    Relative transformation uncertainty is computed with respect to the last predicted pose $\dPose_k$, either by neglecting cross terms and assuming independence~\cite{jung2023ral} (red ellipses) or by considering the correlation between poses as the proposed method (purple ellipses).
    The uncertainties are illustrated as 95\% confidence regions and validated using Monte Carlo simulation with 100 samples (black dots). The proposed method provides a more accurate characterization, while the independence assumption leads to overestimation. Results are shown every 20 IMU inputs for clarity.
    }
    \label{fig:UAMC_analysis}
    \vsfig
\end{figure}
Let a measurement be obtained at $t_k$, where a sample is observed at $\rho_j \in [t_{k-1}, t_k]$. 
We transform the sample using the relative transformation as:
\begin{equation} 
    \dPoseFromFrameKthIMUToJthIMU = \dPoseInvFromFrameWorldToKthIMU \dPoseFromFrameWorldToJthIMU.
    \label{eq:relative_transformation}
\end{equation}
where $\dPoseFromFrameWorldToKthIMU$ and $\dPoseFromFrameWorldToJthIMU$ denote the IMU poses at times $t_k$ and $\rho_j$, respectively and both poses can be obtained from IMU propagation.
Despite advancements in IMU propagation, predicted poses inherently contain errors due to IMU noise and numerical integration, which in turn introduce errors in the relative transformation. 
Such errors significantly contribute to sample deviations, particularly for measurements at distant ranges such as LiDAR point clouds. 
To address this issue, we incorporate the uncertainty of the relative transformation into the sample noise during motion compensation, yielding a probabilistic representation of undistorted measurements.

\begin{figure}[t]
    \captionsetup{font=footnotesize}
    \centering
    \def\twosubfigsize{0.23\textwidth}
    \def\threesubfigsize{0.15\textwidth}
    \begin{subfigure}[b]{\threesubfigsize}
        \centering
        \includegraphics[width=1.0\textwidth]{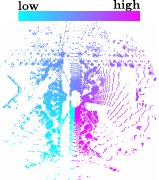}
        \caption{Aquisition time}
    \end{subfigure}
    \begin{subfigure}[b]{\threesubfigsize}
        \centering
        \includegraphics[width=1.0\textwidth]{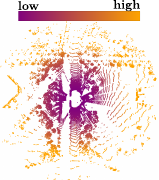}
        \caption{Raw noise}
    \end{subfigure}
    \begin{subfigure}[b]{\threesubfigsize}
        \centering
        \includegraphics[width=1.0\textwidth]{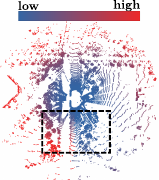}
        \caption{Undistorted noise}
    \end{subfigure}
    \caption{Resultant noise of the undistorted point cloud after applying uncertainty-aware motion compensation (UAMC). 
    (a) Points are acquired clockwise over time.
    (b) The noise of raw measurements increases with distance from the center.
    (c) The point cloud is undistorted through UAMC. 
    As highlighted, points acquired at the beginning exhibit larger noise than those acquired later because earlier points are transformed using relative transformations with higher uncertainty.
    }
    \label{fig:UAMC_result}
    \vsfig
\end{figure}
The true relative transformation, $\dPoseFromFrameKthIMUToJthIMU$, can be expressed using the nominal pose and its perturbation as:
\begin{equation} 
    \begin{aligned}
        \dPoseNomFromFrameKthIMUToJthIMU \dBigExp(\dPosePertFromFrameKthIMUToJthIMU) &= (\dPoseNomFromFrameWorldToKthIMU \dBigExp(\dPosePertFromFrameWorldToKthIMU))^{-1} \dPoseNomFromFrameWorldToJthIMU \dBigExp(\dPosePertFromFrameWorldToJthIMU)\\
        &= \dBigExp(-\dPosePertFromFrameWorldToKthIMU) \dPoseNomInvFromFrameWorldToKthIMU \dPoseNomFromFrameWorldToJthIMU \dBigExp(\dPosePertFromFrameWorldToJthIMU).
    \end{aligned}
    \label{eq:relative_transformation_with_perturbation}
\end{equation}
The perturbations of $\dPoseFromFrameWorldToJthIMU$ and $\dPoseFromFrameWorldToKthIMU$ follow a joint distribution:
\begin{equation} 
    \begin{bmatrix} \dPosePertFromFrameWorldToJthIMU \\ \dPosePertFromFrameWorldToKthIMU \end{bmatrix} \sim \mathcal{N} (\mathbf{0}, \dPoseCov), \quad \dPoseCov = \begin{bmatrix} \dPoseCov_{\dJthIMUFrame} & \dPoseCov_{\dJthIMUFrame, \dKthIMUFrame} \\ \dPoseCov_{\dJthIMUFrame, \dKthIMUFrame}^{\top} & \dPoseCov_{\dKthIMUFrame} \end{bmatrix}
    \label{eq:relationship_between_two_error_state}
\end{equation}
where $\dPoseCov_{\dJthIMUFrame, \dKthIMUFrame}$ is nonzero if the two poses are jointly distributed.
Letting $\dPoseNomFromFrameKthIMUToJthIMU = \dPoseNomInvFromFrameWorldToKthIMU \dPoseNomFromFrameWorldToJthIMU$, we can formulate \eqref{eq:relative_transformation_with_perturbation} as:
\begin{equation} 
    \dBigExp(\dPosePertFromFrameKthIMUToJthIMU) = \dBigExp(\dPosePertPrimeFromFrameWorldToKthIMU) \dBigExp(\dPosePertFromFrameWorldToJthIMU)
    \label{eq:relative_transformation_perturbation_in_exponential_map }
\end{equation}
where $\dPosePertPrimeFromFrameWorldToKthIMU  = -\mathrm{Ad}_{\dPoseNomInvFromFrameKthIMUToIthIMU} \dPosePertFromFrameWorldToKthIMU$ and $\mathrm{Ad}_{\dPose}$ denotes the adjoint matrix of $\dPose$\cite{Mangelson2020tro,Barfoot2014tro}.
Applying the Baker-Campbell-Hausdorff (BCH) formula, the perturbation in the exponential map is expressed as\cite{Barfoot2014tro}:
\begin{equation} 
    \dPosePertFromFrameKthIMUToJthIMU = \dPosePertPrimeFromFrameWorldToKthIMU + \dPosePertFromFrameWorldToJthIMU + \frac{1}{2} \dPosePertPrimeCurwedFromFrameWorldToKthIMU \dPosePertFromFrameWorldToJthIMU
    + \frac{1}{12} \dPosePertPrimeCurwedFromFrameWorldToKthIMU \dPosePertPrimeCurwedFromFrameWorldToKthIMU \dPosePertFromFrameWorldToJthIMU + \cdots . 
    \label{eq:perturbation_of_relative_transformation}
\end{equation}
By taking the expectation of the outer product of \eqref{eq:perturbation_of_relative_transformation}, the relative transformation covariance can be computed as
\begin{equation} 
    \begin{aligned}
        \mathrm{E}[\dPosePertFromFrameKthIMUToJthIMU \dPosePertTransFromFrameKthIMUToJthIMU]  \approx & \underbrace{\mathrm{Ad}_{\dPoseNomInvFromFrameKthIMUToJthIMU} \dPoseCov_{\dKthIMUFrame} \mathrm{Ad}_{\dPoseNomInvFromFrameKthIMUToIthIMU}^{\top} + \dPoseCov_{\dJthIMUFrame}}_{\text{2nd order diag. terms}} \\
        & \underbrace{- \mathrm{Ad}_{\dPoseNomInvFromFrameKthIMUToJthIMU} \dPoseCov_{\dJthIMUFrame, \dKthIMUFrame}^{\top} - \dPoseCov_{\dJthIMUFrame, \dKthIMUFrame} \mathrm{Ad}_{\dPoseNomInvFromFrameKthIMUToJthIMU}^{\top}}_{\text{2nd order cross terms}} + \cdots
    \end{aligned}
    \label{eq:covariance_of_relative_transformation}
\end{equation}
where the cross terms are valid when the poses are jointly distributed as described in \eqref{eq:relationship_between_two_error_state}\cite{Mangelson2020tro}.
In other words, when the predicted poses obtained from IMU propagation are jointly distributed, the cross terms should be considered. Otherwise, they are neglected.

\begin{figure*}[t]
    \captionsetup{font=footnotesize}
    \centering
    \includegraphics[width=0.96\textwidth]{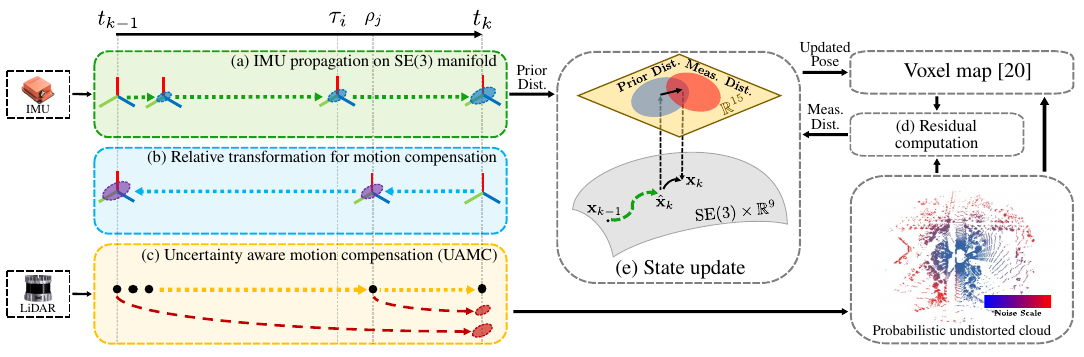}
    \caption{Pipeline of $\dSE3$-LIO that integrates the proposed IMU propagation and uncertainty-aware motion compensation (UAMC).
    Assume the pose at $t_{k-1}$ has been estimated, and a new LiDAR point cloud is acquired at $t_k$.
    (a) IMU propagation is performed on $\dSE3$ manifold using IMU measurements acquired over $[t_{k-1}, t_k]$, providing a prior distribution for the state update.
    (b) Relative transformations, along with their uncertainties, are computed from the reference pose at $t_k$ to poses at each point acquisition time.
    (c) These relative transformations are used to compensate for motion distortion in the raw point cloud, while their uncertainties are incorporated into the noise of each point, resulting in a probabilistic undistorted point cloud.
    (d) The undistorted point cloud is transformed into the world frame and used to construct residuals with the local map, for which we adopt VoxelMap~\cite{Yuan2022ral} to obtain a measurement distribution.
    (e) Finally, the prior and measurement distributions are optimized through the update step, yielding an updated pose estimate at $t_k$.
    }
    \label{fig:framework}
    \vsfig
\end{figure*}

In this study, we derive the joint distribution between the predicted poses and demonstrate that they are jointly distributed.
This property stems from the fact that the IMU propagation is recursively performed, where the current state and an IMU input are used to predict the next state, as \eqref{eq:our_discretized_kinematic_model}.
Under the recursive structure, each error state can be expressed by an initial error state $\dStateErr_0$ and IMU noise $\mathbf{w}$ as:
\begin{equation} 
    \dStateErr_{i+1} = \mathbf{F}_{\dStateErr_{0} \dStateErr_{i}} \dStateErr_{0} + \sum_{\tau =0}^{i} (\mathbf{F}_{\dStateErr_{\tau+1} \dStateErr_{i}} \mathbf{F}_{\mathbf{w}_{\tau}} \mathbf{w})
    \label{eq:error_state_compact_representation}
\end{equation}
where the Jacobians are defined as:
\begin{equation} 
    \mathbf{F}_{\dStateErr_{i} \dStateErr_{k}} =
    \begin{cases}
    \displaystyle\prod_{j=i}^{k-1} \mathbf{F}_{\dStateErr_{j}} & \text{if } k > i \\
    \mathbf{I} & \text{if } k \leq i
    \end{cases}.
    \label{eq:error_state_reformulation_jacobians}
\end{equation}
Stacking the error states over steps, we construct a joint distribution of error states as:
\begin{equation} 
    \setlength{\arraycolsep}{1.5pt} 
    \scriptstyle
    \begin{bmatrix} \dStateErr_{0} \\ \vdots \\ \dStateErr_{i+1} \\ \vdots \\ \dStateErr_{k} \end{bmatrix}  = 
    \begin{bmatrix} 
        \dI & \dZero & \cdots & \dZero & \cdots & \dZero \\
        \vdots & \vdots & \ddots & \vdots & \ddots & \vdots \\
        \mathbf{F}_{\dStateErr_{0} \dStateErr_{i}} & \mathbf{F}_{\dStateErr_{1}\dStateErr_{i}} & \cdots & \dI & \cdots & \dZero \\
        \vdots & \vdots & \ddots & \vdots & \ddots & \vdots \\
        \mathbf{F}_{\dStateErr_{0} \dStateErr_{k-1}} & \mathbf{F}_{\dStateErr_{1}\dStateErr_{k-1}} & \cdots & \mathbf{F}_{\dStateErr_{i+1}\dStateErr_{k-1}} & \cdots & \dI
    \end{bmatrix}
    \begin{bmatrix} \dStateErr_{0} \\  \mathbf{w}_{0}^{'}  \\ \vdots \\ \mathbf{w}_{i}^{'} \\ \vdots \\ \mathbf{w}_{k-1}^{'} \end{bmatrix}.
    \label{eq:joint_error_state}
\end{equation}
where $\mathbf{w}_{i}^{'} = \mathbf{F}_{\mathbf{w}_{i}} \mathbf{w}$.
By computing the expectation of outer product of \eqref{eq:joint_error_state}, we obtain the joint covariance of error states, which contains the joint covariance of pose described in \eqref{eq:relationship_between_two_error_state}.
This enables accurate characterization of the relative transformation covariance using \eqref{eq:covariance_of_relative_transformation} by incorporating the cross terms.

To the best of our knowledge, MA-LIO~\cite{jung2023ral} is the first method to incorporate the relative transformation uncertainty into measurement noise during motion compensation.
However, it neglects the cross terms in \eqref{eq:covariance_of_relative_transformation} by assuming that the predicted poses are independent\cite{Barfoot2014tro}.
To analyze the effect of considering cross terms in uncertainty characterization, we conducted an experiment similar to the one proposed in \cite{Mangelson2020tro} as shown in \figref{fig:UAMC_analysis}.
The results demonstrate that the proposed method characterizes the relative transformation uncertainty more accurately, while the independent assumption leads to overestimation.
Additionally, in the proposed method, the uncertainty increases as the target pose lies further in the past with respect to the reference pose.
In contrast, the independent assumption exhibits a decreasing trend opposite to the proposed method because the earlier target pose $\dPoseCov_{\dJthIMUFrame}$ in \eqref{eq:covariance_of_relative_transformation} has a smaller uncertainty.


Let's consider a distorted LiDAR point $\dJthPtInJthLidarFrame$, which is acquired at time $\rho_j$ and represented in the LiDAR frame.
The point is undistorted using the relative transformation as: 
\begin{equation} 
    \dJthPtHomoInKthIMUFrame = \dPoseFromFrameKthIMUToJthIMU \dPoseFromFrameIMUToLidar \dJthPtHomoInJthLidarFrame,
    \quad \tau_i \leq \rho_j < \tau_{i+1}
    \label{eq:motion_compensation}
\end{equation}
where $\dPoseFromFrameIMUToLidar$ is the extrinsic transformation from IMU to LiDAR frame and $\dPtHomo$ is a point in homogeneous coordinates.
The undistorted point is modeled as:
\begin{equation} 
    \dJthPtHomoInKthIMUFrame = \dJthPtNomHomoInKthIMUFrame + \dJthPtNoiseDilInKthIMUFrame
    = \dPoseNomFromFrameKthIMUToJthIMU \dBigExp(\dPosePertFromFrameKthIMUToJthIMU) (\dJthPtNomHomoInJthIMUFrame + \dJthPtNoiseDilInJthIMUFrame)
    \label{eq:undistorted_point_with_noise}
\end{equation}
where the true point $\dPtHomo$ consists of a nominal point $\dPtNomHomo$ and a noise term $\dPtNoiseDil$.
The noise is expressed in homogeneous coordinates using a dilation matrix $\mathbf{D}$ that maps a 3D vector to a homogeneous form, such that $\dPtNoiseDil = \mathbf{D} \mathbf{n}$~\cite{Jiao2022tro}.
Letting $\dJthPtNomHomoInKthIMUFrame = \dPoseNomFromFrameKthIMUToJthIMU \dJthPtNomHomoInJthIMUFrame$, and by linearizing \eqref{eq:undistorted_point_with_noise}, the noise term $\dJthPtNoiseDilInKthIMUFrame$ can be approximated as:
\begin{equation} 
    \dJthPtNoiseDilInKthIMUFrame \approx \dPoseNomFromFrameKthIMUToJthIMU (\dJthPtNomHomoInJthIMUFrame)^{\odot} \dPosePertFromFrameKthIMUToJthIMU + \dPoseNomFromFrameKthIMUToJthIMU \dJthPtNoiseDilInJthIMUFrame.
    \label{eq:linearlized_undistorted_point_with_noise}
\end{equation}
where $(\cdot)^{\odot}$ denotes an operator that maps a 4$\times$1 vector to a 4$\times$6 matrix\cite{Jiao2022tro, Yuan2022ral}.
The resulting noise $\dJthPtNoiseDilInKthIMUFrame$ reflects both the uncertainty of the relative transformation and the raw measurement noise, as illustrated in \figref{fig:UAMC_result}.
\subsection{Integration with LiDAR-Inertial Odometry}
\label{subsec:integration_with_LIO}

We propose an error-state Kalman filter-based LiDAR-inertial odometry (LIO) framework, termed $\dSE3$-LIO, that integrates the proposed smooth IMU propagation method and UAMC.
The framework optimizes the prior obtained from the proposed IMU propagation method from \secref{subsec:imu_propagation_on_SE3_manifold} and the measurement distribution constructed by a probabilistic undistorted point cloud from \secref{subsec:UAMC}.
Pipeline of $\dSE3$-LIO is illustrated in \figref{fig:framework}.

IMU propagation is performed until the end of a LiDAR scan at $t_{k}$, resulting in a prior state as:
\begin{equation} 
    \dKthState = \dKthStateNom \boxplus \dKthStateErr, \quad \dKthStateErr \sim \mathcal{N}(\dZero, \hat{\mathbf{P}}_{k}).
    \label{eq:prior_state}
\end{equation}
Then the probabilistic undistorted point cloud obtained through UAMC is transformed to the world frame to construct residuals with the local map, for which we adopt voxel map~\cite{Yuan2022ral} as:
\begin{equation} 
    \mathbf{h}_j(\dKthState, \dJthPtInKthIMUFrame) = (\mathbf{u}_{j}^{\mathrm{d}})^{\top} (\dPoseFromFrameWorldToKthIMU \dJthPtHomoInKthIMUFrame - \mathbf{q}_{j}^{\mathrm{d}})
    \label{eq:residual}
\end{equation} 
where $\mathbf{u}_{j}^{\mathrm{d}}$, $\mathbf{q}_{j}^{\mathrm{d}}$ are the normal vector and the center of the plane, respectively, in homogeneous coordinates.
The residual can be linearized as:
\begin{equation} 
    \mathbf{h}_j(\dKthState, \dJthPtInKthIMUFrame) \approx \mathbf{h}_j(\dKthStateNom, \dJthPtNomInKthIMUFrame) + \mathbf{H}_j \dKthStateErr + \mathbf{J}_j \dJthPtNoiseDilInKthIMUFrame
    \label{eq:residual_linearization}
\end{equation}
where $\mathbf{H}_j$ and $\mathbf{J}_j$, are the Jacobians of $\mathbf{h}_j$ with respect to $\dKthState$ and $\dJthPtNoiseDilInKthIMUFrame$, respectively, evaluated at $\dKthState = \dKthStateNom$ and $\dJthPtInKthIMUFrame = \dJthPtNomInKthIMUFrame$ as:
\begin{equation} 
    \mathbf{H}_j = \begin{bmatrix} (\mathbf{u}_{j}^{\mathrm{d}})^{\top} (\dJthPtNomHomoInKthIMUFrame)^{\odot} \dPoseNomFromFrameKthIMUToJthIMU & \dZero_{1 \times 12} \end{bmatrix}, \;
    \mathbf{J}_j = (\mathbf{u}_{j}^{\mathrm{d}})^{\top} \dPoseNomFromFrameKthIMUToJthIMU.
    \label{eq:residual_jacobians}
\end{equation}
Combining the prior from \eqref{eq:prior_state} and the measurement distribution from \eqref{eq:residual_linearization} yields the maximum a posteriori (MAP) estimate as:
\begin{equation} 
    \underset{\dKthStateErr}{\mathrm{min}} \; ( \left\| \dKthStateErr \right\|^2_{\hat{\mathbf{P}}_{k}^{-1}}  +
    \sum_{j}^{m} \left\| \mathbf{z}_{j} + \mathbf{H}_{j} \dKthStateErr \right\|^2_{\mathbf{R}_{j}^{-1}} ) 
    \label{eq:MAP_estimate}
\end{equation}
where $\left\| \dState \right\|_{\mathbf{M}} = \dStateTrans \mathbf{M} \dState$.
Standard Kalman filter update step is performed by optimizing the resultant quadratic cost\cite{wei2021ral, Yuan2022ral}.

\section{Experimental results}
\label{sec:exp}
We perform an extensive evaluation on $\dSE3$-LIO across diverse scenarios.
Experiments are conducted to validate the effectiveness of $\dSE3$-LIO in terms of accuracy and robustness in challenging environments, as well as computational efficiency compared with state-of-the-art LIO methods.

\subsection{Experiment Setup}
\label{subsec:experiment_setup}

\begin{figure}[t]
    \centering
    \includegraphics[width=0.48\textwidth]{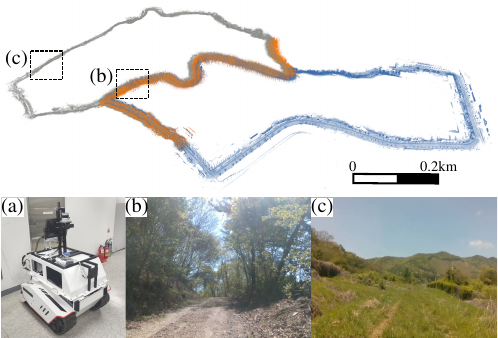}
    \captionsetup{font=footnotesize}
    \caption{
        Description of our dataset, including $\texttt{CW}$ (blue), $\texttt{CCW}$ (gray), and $\texttt{mid}$ (orange) sequences.
        (a) A ground vehicle equipped with a 128-channel LiDAR (Ouster OS2-128), an IMU (Xsens MTi-300), and a RTK-GNSS receiver (Emlid Reach RS).
        (b)\textendash(c) Data were collected in large-scale rough terrain, including dense forest and wide open areas.
    } 
    \label{fig:boeun_map}
    \vsfig
\end{figure}

We conduct experiments on three datasets, including NTU-VIRAL~\cite{nguyen2022ijrr}, Newer-College~\cite{ramezani2020iros}, and our own dataset.
The NTU-VIRAL and Newer-College datasets are utilized to evaluate performance under challenging conditions, including aggressive motion and sparse measurements.
The NTU-VIRAL dataset was collected using a drone equipped with a low-channel LiDAR (Ouster OS1-16) and features aggressive motion with rapid directional changes and high linear acceleration.
The Newer-College dataset was collected using a handheld platform, where dynamic motion was induced by human operators shaking it while walking.
Finally, our own dataset was collected using a ground vehicle in large-scale rough terrain, featuring wild fluctuations due to uneven terrain and degraded measurements due to dense vegetation and extensive open spaces, as shown in \figref{fig:boeun_map}.
We obtain ground truth poses for our dataset using a RTK-GNSS system (Emlid Reach RS) mounted on the vehicle.

$\dSE3$-LIO is compared with state-of-the-art LiDAR-inertial odometry methods, including FAST-LIO2~\cite{wei2022tro}, PV-LIO\footnote{https://github.com/HViktorTsoi/PV-LIO}, DLIO~\cite{chen2023icra}, LIO-EKF~\cite{wu2024icra}, and MA-LIO~\cite{jung2023ral}.
PV-LIO is extended from VoxelMap~\cite{Yuan2021ral} by integrating IMU propagation.
For a fair comparison, we adopted the default parameter set provided in the original works for all baseline methods, and we applied an identical parameter set for $\dSE3$-LIO within each dataset.
All methods were tested with a 32-core Intel i9-14900K CPU.
\begin{table*}[t!]
    \captionsetup{font=footnotesize}
    \centering
    \caption{The root mean square error (RMSE) of absolute trajectory error (ATE) (m) on the NTU, Newer-College, and our datasets. \textbf{Bold} value denotes the best performance, \underline{underline} value denotes the second-best performance and \textbf{‘-’} denotes that the odometry has diverged.}
    {
        \begin{tabular}{l|ccccccccccccc}
            \toprule 
            Dataset & \multicolumn{6}{c}{NTU-VIRAL\cite{nguyen2022ijrr}} & \multicolumn{4}{c}{Newer-College\cite{ramezani2020iros}} & \multicolumn{3}{c}{Our dataset} \\
             \cmidrule(lr){2-7} \cmidrule(lr){8-11} \cmidrule(lr){12-14}
            Sequence    & $\texttt{eee}$ & $\texttt{nya}$ & $\texttt{rtp}$ & $\texttt{sbs}$ & $\texttt{tnp}$ & $\texttt{spms}$ & $\texttt{quad-D}$ & $\texttt{quad-M}$ & $\texttt{quad-H}$ & $\texttt{math-H}$ & $\texttt{CW}$ & $\texttt{CCW}$ & $\texttt{mid}$ \\
            \midrule
            FAST-LIO2\cite{wei2022tro}   & \underline{0.250} & 0.256 & 0.276 & \underline{0.27} & \underline{0.244} & 3.12 & 0.356 & 0.064 & \textbf{0.084} & \underline{0.084} & 14.07 & \underline{17.08} & 2.50 \\
            PV-LIO      & 0.579 & 0.404 & 1.71 & 0.753 & 0.74 & - & 0.135 & 0.093 & 0.107 & \textbf{0.082} & 13.47 & 21.62 & \textbf{2.00} \\
            DLIO\cite{chen2023icra}      & 0.254 & 0.269 & \underline{0.231} & 0.286 & 0.258 & \underline{0.423} & \underline{0.150} & 0.076 & 2.66 & 0.116 & \underline{12.34} & 43.04 & 3.77 \\
            LIO-EKF\cite{wu2024icra}     & 0.772 & 0.479 & 1.04 & 0.308 & 0.365 & 2.45 & 2.48 & 0.314 & 2.95 & 1.21 & - & 40.35 & - \\
            MA-LIO\cite{jung2023ral}      & 0.252 & \underline{0.256} & 2.32 & \textbf{0.268} & 0.248 & 0.464 & 0.156 & \underline{0.060} & 0.307 & 0.102 & 76.82 & 27.77 & 3.09 \\
            $\dSE3$-LIO (ours)        & \textbf{0.246} & \textbf{0.253} & \textbf{0.204} & \textbf{0.268} & \textbf{0.238} & \textbf{0.321} & \textbf{0.128} & \textbf{0.056} & \underline{0.099} & \textbf{0.082} & \textbf{5.10} & \textbf{15.86} & \underline{2.29} \\
            \bottomrule
        \end{tabular}
    }
    \vspace{0.1cm}
    \label{tab:quantitative_results}
\end{table*}

\begin{figure*}[t]
    \captionsetup{font=footnotesize}
    \centering
    \includegraphics[width=0.96\textwidth]{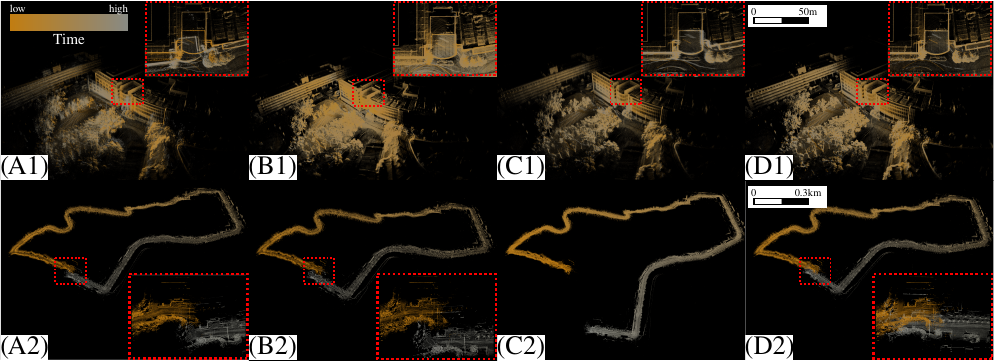}
    \caption{Qualitative comparison in $\texttt{spms}$ sequence of the NTU-VIRAL\cite{nguyen2022ijrr} dataset (upper) and $\texttt{CW}$ sequence of our dataset (lower). 
    Baselines including (A) Fast-LIO2\cite{wei2022tro}, (B) DLIO\cite{chen2023icra}, and (C) MA-LIO\cite{jung2023ral} are compared with (D) $\dSE3$-LIO (ours).
Point clouds are colored by time, where brown and gray indicate the beginning and the end of the sequence, respectively.
    Highlighted boxes indicate the top-down view of the corresponding areas, showing the drift in estimated trajectory.
    Resultant point clouds are captured by identical viewpoints for fair comparison.
    }
    \label{fig:qualitative_result}
    \vsfig
\end{figure*}

\subsection{Quantitative Results}
\label{subsec:quantitative_results}

In the first experiment, we evaluate the accuracy of $\dSE3$-LIO on three datasets, as presented in \tabref{tab:quantitative_results}.
As a quantitative metric, we use root mean square error (RMSE) of absolute trajectory error (ATE) using the open-source evaluation toolkit\footnote{https://github.com/MichaelGrupp/evo}.
In the NTU-VIRAL dataset, $\dSE3$-LIO achieves the most accurate results across all sequences.
While MA-LIO~\cite{jung2023ral} and DLIO~\cite{chen2023icra}, which address aggressive motion through improved IMU propagation, exhibit comparable performance in most sequences,
both methods still degenerate in $\texttt{spms}$ sequence, where the rotational and translational speeds exceed 1.6 $\mathrm{rad/s}$ and 5.0 $\mathrm{m/s}$, respectively.
Moreover, $\dSE3$-LIO and DLIO demonstrate small deviations in results across all sequences, indicating robustness against varying dynamic motions.
In the Newer-College dataset, we manually downsample the point clouds with a voxel size of 2 $\mathrm{m}$ to further increase measurement sparsity, and $\dSE3$-LIO achieves the best performance across all sequences, except for the quad-H sequence where it exhibits second-best performance.


In our dataset, $\dSE3$-LIO achieves superior performance in large-scale rough terrain, especially in $\texttt{CW}$ and $\texttt{CCW}$ sequences, where the robot traveled a total distance of 2.4 and 2.7 $\mathrm{km}$, respectively.
Overall errors are lower in $\texttt{CW}$ than in $\texttt{CCW}$, since $\texttt{CW}$ includes both structured and unstructured environments, whereas $\texttt{CCW}$ mainly consists of unstructured environments.
Notably, $\dSE3$-LIO reduces error in the $\texttt{CW}$ sequence by more than half compared with the second-best method, demonstrating accuracy against wild fluctuations and degraded measurements.

We also conduct an ablation study to validate the contributions of each proposed component, as shown in \tabref{tab:ablation_study}.
Performing IMU propagation on $\dSE3$ manifold instead of $\dSO3 \times \dR3$ yields substantial improvements across all sequences, especially in aggressive motion sequences, such as $\texttt{rtp}$ and $\texttt{spms}$.
Additionally, incorporating UAMC consistently improves accuracy, validating its contribution to the overall performance.
\subsection{Qualitative Results}
\label{sec:qualitative_results}

We present qualitative comparisons on $\texttt{spms}$ from NTU-VIRAL dataset and $\texttt{CW}$ from our dataset in \figref{fig:qualitative_result}.
Both sequences exhibit loop trajectories in which odometry returns to origin.
Point clouds are colored by time, allowing drift in estimated trajectory to be clearly identified.
In $\texttt{spms}$, FAST-LIO2~\cite{wei2022tro} exhibits significant drift under aggressive motion, such as rapid rotations and high accelerations.
While DLIO~\cite{chen2023icra} and MA-LIO~\cite{jung2023ral} achieve reliable results thanks to their advanced IMU propagation methods, they still suffer from drift due to sparse measurements from low-channel LiDAR.
In contrast, $\dSE3$-LIO exhibits the least drift by providing an accurate prior through smooth IMU propagation and a measurement distribution via UAMC.
In $\texttt{CW}$, all methods show noticeable drift, except for $\dSE3$-LIO.
MA-LIO~\cite{jung2023ral} fails to estimate poses in the middle of the trajectory due to degraded measurements in unstructured environments.
However, $\dSE3$-LIO maintains robust and consistent performance throughout the entire trajectory.
\begin{table}[t]
    \vspace{4pt}
    \captionsetup{font=footnotesize}
    \centering
    \caption{Ablation study on NTU-VIRAL~\cite{nguyen2022ijrr} dataset. RMSE of ATE (m) is reported. \textbf{Bold} denotes the best performance.}
    \setlength{\tabcolsep}{4.5pt}
    {\scriptsize
        \begin{tabular}{l|cccccc}
            \toprule
            Sequences &$\texttt{eee}$ & $\texttt{nya}$ & $\texttt{rtp}$ & $\texttt{sbs}$  & $\texttt{tnp}$ & $\texttt{spms}$ \\
            \midrule
            $\dSO3 \times \dR3$ (baseline)   & 0.548 & 0.469 & 2.64 & 0.750  & 0.693 & 0.638 \\
            $\dSE3$ w/o UAMC      & 0.324 & 0.259 & \textbf{0.268} & 0.258  & 0.216 & 0.241 \\
            $\dSE3$ w/ UAMC  & \textbf{0.321} & \textbf{0.246} & 0.269 &  \textbf{0.253}  & \textbf{0.204} & \textbf{0.238} \\
            \bottomrule
        \end{tabular}
    }
    \label{tab:ablation_study}
\end{table}
\subsection{Computation Time Evaluation}
\label{sec:time_usage_evaluation}

\begin{figure}[t]
    \centering
    \includegraphics[width=0.48\textwidth]{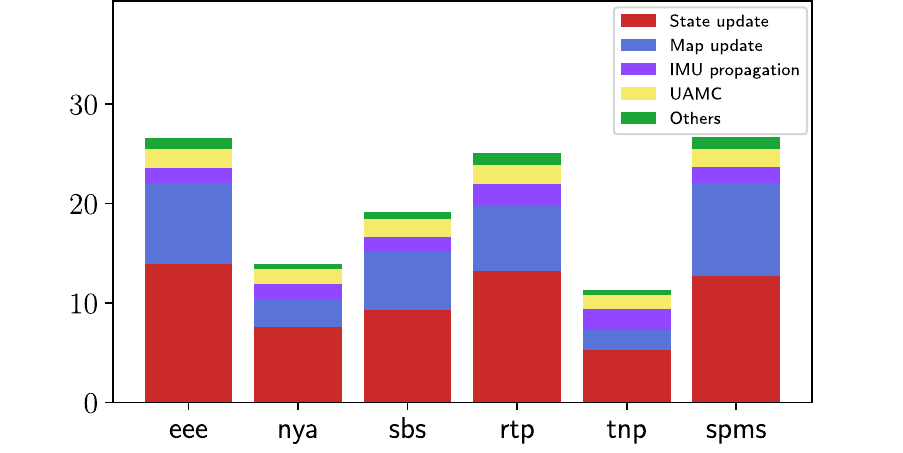}
    \captionsetup{font=footnotesize}
    \caption{
        Time usage for each module of the proposed LIO framework in NTU-VIRAL~\cite{nguyen2022ijrr} dataset.
    }
    \label{fig:time_evaluation}
    \vsfig
\end{figure}

We evaluate the average computation time of the baselines and the proposed method in \tabref{tab:time_evaluation}.
FAST-LIO2~\cite{wei2022tro} achieves the lowest computation time owing to its efficient framework, whereas DLIO~\cite{chen2023icra} exhibits the highest computation time due to its advanced motion compensation process.
$\dSE3$-LIO is on par with FAST-LIO2~\cite{wei2022tro}, while yielding superior performance.
Furthermore, we analyze the computation time of each module in $\dSE3$-LIO as illustrated in \figref{fig:time_evaluation}.
The proposed IMU propagation and UAMC account for only 9.6\% and 8.9\% of the total computation time, respectively, indicating that they introduce minimal computational overheads to the overall system.
The majority of the computation time is attributed to state and map updates, which follow the implementations of FAST-LIO2 and VoxelMap\cite{Yuan2022ral}.

\section{Conclusion}
\label{sec:conclusion}


In this study, we presented a smooth IMU propagation method with jointly distributed poses on SE(3) manifold, addressing the challenges of motion prediction and motion compensation in existing IMU propagation methods.
Furthermore, we proposed SE(3)-LIO that integrates the proposed IMU propagation.
Experimental results demonstrate that SE(3)-LIO outperforms state-of-the-art LIO methods in terms of accuracy and robustness.
Despite these promising results, there is room for improvements in terms of adaptability to different sensor configurations.
As future work, we plan to extend the proposed IMU propagation method to other sensor configurations, such as visual-inertial or radar\cite{lim2023icra} odometry.

\begin{table}[t]
    \captionsetup{font=footnotesize}
    \centering
    \caption{Comparison of average computation time (ms) of baselines and ours in NTU-VIRAL~\cite{nguyen2022ijrr} dataset.
    \textbf{Bold} denotes the lowest computation time, and \underline{underline} denotes the second-lowest computation time.
    }
    \setlength{\tabcolsep}{4.5pt}
    {\scriptsize
        \begin{tabular}{l|ccccccc}
            \toprule
            Sequences & $\texttt{eee}$ & $\texttt{nya}$ & $\texttt{sbs}$ & $\texttt{rtp}$ & $\texttt{tnp}$ & $\texttt{spms}$ & avg \\
            \midrule
            FAST-LIO2\cite{wei2022tro} &  \textbf{16.02} & \underline{14.15} & \textbf{18.44} & \textbf{14.24} & \underline{14.80} & \textbf{17.74} & \textbf{15.90} \\
            PV-LIO\cite{ramezani2020iros} &  51.84 & 38.92 & 46.49 & 49.79 & 36.48 & 44.87 & 44.73 \\
            DLIO\cite{chen2023icra} &  99.08 & 98.60 & 100.21 & 98.89 & 99.78 & 100.44 & 99.50 \\
            MA-LIO\cite{jung2023ral} &  \underline{23.97} & 23.77 & 28.52 & 23.33 & 24.22 & \underline{26.38} & 25.03 \\
            $\dSE3$-LIO (ours) &  26.66 & \textbf{13.95} & \underline{25.03} & \underline{19.15} & \textbf{11.32} & 26.66 &  \underline{20.46} \\    
            \bottomrule
        \end{tabular}
    } 
    \label{tab:time_evaluation}
\end{table}


\bibliographystyle{URL-IEEEtrans}

\bibliography{URL-bib}

@STRING{ral  = {IEEE Robot. Automat. Lett.} }

@STRING{tro  = {IEEE Trans. Robot.} }

@STRING{ijrr = {Int. J. Robot. Res.} }

@STRING{isr  = {Intell. Serv. Robot.} }

@STRING{rss    = {Robot. Sci. Syst.} }

@STRING{iros   = {Proc. IEEE/RSJ Int. Conf. Intell. Robot. Syst.} }

@STRING{icra   = {Proc. IEEE Int. Conf. Robot. Automat.} }

@STRING{ur     = {Proc. Int. Conf. Ubiquti. Robot.} }

@article{wei2021ral,
  author  = {Xu, Wei and Zhang, Fu},
  journal = ral,
  title   = {{FAST-LIO}: A Fast, Robust {LiDAR}-Inertial Odometry Package by Tightly-Coupled Iterated {Kalman} Filter},
  year    = {2021},
  volume  = {6},
  number  = {2},
  pages   = {3317-3324}
}

@article{wei2022tro,
  title   = {{FAST-LIO2}: Fast direct {LiDAR}-inertial odometry},
  author  = {Xu, Wei and Cai, Yixi and He, Dongjiao and Lin, Jiarong and Zhang, Fu},
  journal = tro,
  volume  = {38},
  number  = {4},
  pages   = {2053--2073},
  year    = {2022}
}

@article{Mangelson2020tro,
  author  = {Mangelson, Joshua G. and Ghaffari, Maani and Vasudevan, Ram and Eustice, Ryan M.},
  journal = tro,
  title   = {Characterizing the Uncertainty of Jointly Distributed Poses in the {Lie} Algebra},
  year    = {2020},
  volume  = {36},
  number  = {5},
  pages   = {1371-1388}
}

@article{Yuan2021ral,
  author  = {Yuan, Chongjian and Liu, Xiyuan and Hong, Xiaoping and Zhang, Fu},
  journal = ral,
  title   = {Pixel-Level Extrinsic Self Calibration of High Resolution {LiDAR} and Camera in Targetless Environments},
  year    = {2021},
  volume  = {6},
  number  = {4},
  pages   = {7517-7524}
}

@article{Barfoot2014tro,
  author  = {Barfoot, Timothy D. and Furgale, Paul T.},
  journal = tro,
  title   = {Associating Uncertainty With Three-Dimensional Poses for Use in Estimation Problems},
  year    = {2014},
  volume  = {30},
  number  = {3},
  pages   = {679-693}
}

@article{Jiao2022tro,
  author  = {Jiao, Jianhao and Ye, Haoyang and Zhu, Yilong and Liu, Ming},
  journal = tro,
  title   = {Robust Odometry and Mapping for Multi-{LiDAR} Systems With Online Extrinsic Calibration},
  year    = {2022},
  volume  = {38},
  number  = {1},
  pages   = {351-371}
}

@article{jung2023ral,
  author  = {Jung, Minhyuk and Jung, Sungjin and Kim, Ayoung},
  journal = ral,
  title   = {Asynchronous Multiple {LiDAR}-Inertial Odometry Using Point-Wise Inter-{LiDAR} Uncertainty Propagation},
  year    = {2023},
  volume  = {8},
  number  = {7},
  pages   = {4211-4218}
}

@article{Yuan2022ral,
  author  = {Yuan, Chongjian and Xu, Wei and Liu, Xiyuan and Hong, Xiaoping and Zhang, Fu},
  journal = ral,
  title   = {Efficient and Probabilistic Adaptive Voxel Mapping for Accurate Online {LiDAR} Odometry},
  year    = {2022},
  volume  = {7},
  number  = {3},
  pages   = {8518-8525}
}

@article{nguyen2022ijrr,
  title     = {{NTU VIRAL}: A Visual-Inertial-Ranging-{LiDAR} Dataset, From an Aerial Vehicle Viewpoint},
  author    = {Nguyen, {Thien-Minh} and Yuan, Shenghai and Cao, Muqing and Lyu, Yang and Nguyen, Thien Hoang and Xie, Lihua},
  journal   = ijrr,
  volume    = {41},
  number    = {3},
  pages     = {270-280},
  year      = {2022},
  publisher = {SAGE Publications Sage UK: London, England}
}

@article{forster2017tro,
  author={Forster, Christian and Carlone, Luca and Dellaert, Frank and Scaramuzza, Davide},
  journal=tro, 
  title={On-Manifold Preintegration for Real-Time Visual--Inertial Odometry}, 
  year={2017},
  volume={33},
  number={1},
  pages={1-21},
  doi={10.1109/TRO.2016.2597321}
  }

@article{brossard2022tro,
  author={Brossard, Martin and Barrau, Axel and Chauchat, Paul and Bonnabel, Silvére},
  journal=tro, 
  title={Associating Uncertainty to Extended Poses for on {Lie} Group {IMU} Preintegration With Rotating Earth}, 
  year={2022},
  volume={38},
  number={2},
  pages={998-1015},
  doi={10.1109/TRO.2021.3100156}
}

@article{foehn2021sr,
author = {Philipp Foehn  and Angel Romero  and Davide Scaramuzza },
title = {Time-optimal planning for quadrotor waypoint flight},
journal = {Sci. Robot.},
volume = {6},
number = {56},
pages = {eabh1221},
year = {2021},
doi = {10.1126/scirobotics.abh1221}
}

@INPROCEEDINGS{wu2024icra,
  author={Wu, Yibin and Guadagnino, Tiziano and Wiesmann, Louis and Klingbeil, Lasse and Stachniss, Cyrill and Kuhlmann, Heiner},
  booktitle=icra,
  title={{LIO-EKF}: High Frequency {LiDAR}-Inertial Odometry using Extended {Kalman} Filters}, 
  year={2024},
  volume={},
  number={},
  pages={13741-13747}
}

@INPROCEEDINGS{chen2023icra,
  author={Chen, Kenny and Nemiroff, Ryan and Lopez, Brett T.},
  booktitle=icra,
  title={Direct {LiDAR}-Inertial Odometry: Lightweight {LIO} with Continuous-Time Motion Correction}, 
  year={2023},
  volume={},
  number={},
  pages={3983-3989}
}

@ARTICLE{mueggler2018tro,
  author={Mueggler, Elias and Gallego, Guillermo and Rebecq, Henri and Scaramuzza, Davide},
  journal=tro,
  title={Continuous-Time Visual-Inertial Odometry for Event Cameras}, 
  year={2018},
  volume={34},
  number={6},
  pages={1425-1440}
}

@book{murray1994manipulation,
  title     = {A Mathematical Introduction to Robotic Manipulation},
  author    = {R. M. Murray and Z. Li and S. Sastry},
  publisher = {CRC Press},
  address   = {Boca Raton, FL, USA},
  year      = {1994}
}

@INPROCEEDINGS{ramezani2020iros,
  author={Ramezani, Milad and Wang, Yiduo and Camurri, Marco and Wisth, David and Mattamala, Matias and Fallon, Maurice},
  booktitle=iros,
  title={The Newer College Dataset: Handheld {LiDAR}, Inertial and Vision with Ground Truth}, 
  year={2020},
  volume={},
  number={},
  pages={4353-4360},
  doi={10.1109/IROS45743.2020.9340849}
  }

@inproceedings{shan2020iros,
  title={{LIO-SAM}: Tightly-coupled {LiDAR} Inertial Odometry via Smoothing and Mapping},
  author={Shan, Tixiao and Englot, Brendan and Meyers, Drew and Wang, Wei and Ratti, Carlo and Rus Daniela},
  booktitle=iros,
  pages={5135-5142},
  year={2020}
}

@article{ramezani2022arxiv,
  title={Wildcat: Online continuous-time {3D} {LiDAR}-inertial {SLAM}},
  author={Ramezani, Milad and Khosoussi, Kasra and Catt, Gavin and Moghadam, Peyman and Williams, Jason and Borges, Paulo and Pauling, Fred and Kottege, Navinda},
  journal={arXiv preprint arXiv:2205.12595},
  year={2022}
}

@ARTICLE{nguyen2024ral,
  author={Nguyen, Thien-Minh and Xu, Xinhang and Jin, Tongxing and Yang, Yizhuo and Li, Jianping and Yuan, Shenghai and Xie, Lihua},
  journal=ral,
  title={Eigen Is All You Need: Efficient {LiDAR}-Inertial Continuous-Time Odometry With Internal Association}, 
  year={2024},
  volume={9},
  number={6},
  pages={5330-5337}
}

@ARTICLE{qin2018tro,
  author={Qin, Tong and Li, Peiliang and Shen, Shaojie},
  journal=tro, 
  title={{VINS-Mono}: A Robust and Versatile Monocular Visual-Inertial State Estimator}, 
  year={2018},
  volume={34},
  number={4},
  pages={1004-1020}
}

@INPROCEEDINGS{geneva2020icra,
  author={Geneva, Patrick and Eckenhoff, Kevin and Lee, Woosik and Yang, Yulin and Huang, Guoquan},
  booktitle=icra,
  title={{OpenVINS}: A Research Platform for Visual-Inertial Estimation}, 
  year={2020},
  volume={},
  number={},
  pages={4666-4672}
}

@ARTICLE{lee2024isr,
  author={Lee, Dongjae and Jung, Minwoo and Yang, Wooseong and Kim, Ayoung},
  title={{LiDAR} odometry survey: recent advancements and remaining challenges},
  journal=isr,
  year={2024},
  volume={17},
  number={2},
  pages={95-118},
  doi={10.1007/s11370-024-00515-8}
}

@INPROCEEDINGS{qin2020icra,
  author={Qin, Chao and Ye, Haoyang and Pranata, Christian E. and Han, Jun and Zhang, Shuyang and Liu, Ming},
  booktitle=icra,
  title={{LINS}: A {LiDAR}-Inertial State Estimator for Robust and Efficient Navigation}, 
  year={2020},
  volume={},
  number={},
  pages={8899-8906},
  doi={10.1109/ICRA40945.2020.9197567}
}

@INPROCEEDINGS{lim2023ur,
  author={Lim, Hyungtae and Kim, Daebeom and Kim, Beomsoo and Myung, Hyun},
  booktitle=ur,
  title={{AdaLIO}: Robust Adaptive {LiDAR}-Inertial Odometry in Degenerate Indoor Environments}, 
  year={2023},
  volume={},
  number={},
  pages={48-53},
  doi={10.1109/UR57808.2023.10202252}
}

@inproceedings{forster2015rss,
  author = {Forster, Christian and Carlone, Luca and Dellaert, Frank and Scaramuzza, Davide},
  booktitle = rss,
  title = {{IMU} Preintegration on Manifold for Efficient Visual-Inertial Maximum-a-Posteriori Estimation.},
  year = 2015
}

@article{lim2022uv,
  title={{UV-SLAM}: Unconstrained Line-based {SLAM} Using Vanishing Points for Structural Mapping},
  author={Lim, Hyunjun and Jeon, Jinwoo and Myung, Hyun},
  journal=ral,
  year={2022},
  publisher={IEEE},
  volume={7},
  number={2},
  pages={1518-1525},
  doi={10.1109/LRA.2022.3140816}
}

@INPROCEEDINGS{lim2023icra,
  author={Lim, Hyungtae and Han, Kawon and Shin, Gunhee and Kim, Giseop and Hong, Songcheol and Myung, Hyun},
  booktitle=icra,
  title={{ORORA}: Outlier-Robust Radar Odometry}, 
  year={2023},
  volume={},
  number={},
  pages={2046-2053},
  doi={10.1109/ICRA48891.2023.10160997}}

\end{document}